\documentclass[letterpaper, 10 pt, conference]{ieeeconf}  

\makeatletter
\let\NAT@parse\undefined
\makeatother

\usepackage{graphics} 
\usepackage{epsfig} 
\usepackage{times} 
\usepackage{amsmath} 
\usepackage{amssymb}  
\usepackage{subcaption}
\usepackage{xspace}
\usepackage{booktabs}
\usepackage{cite}
\usepackage{caption}

\usepackage{tabularray}
\usepackage{booktabs,multirow}

\usepackage[bookmarks=false,pagebackref=true,breaklinks=true,colorlinks=true,linkcolor=brown,urlcolor=cyan,citecolor=brown]{hyperref}
\usepackage[dvipsnames]{xcolor}
\usepackage{cleveref}
\usepackage{cuted}
\usepackage{url}
\usepackage{comment}
\usepackage{array}

\newif\ifdraft
\drafttrue 
\ifdraft
\usepackage[colorinlistoftodos]{todonotes}
\else
\usepackage[disable]{todonotes}
\fi

\newcommand{\basepolicy}{\pi^\text{base}}
\newcommand{\residualpolicy}{\pi^\text{adapter}}
\newcommand{\privobs}{\mathbf{s}^\text{priv}_t}
\newcommand{\objobs}{\mathbf{s}^\text{obj}_t}
\newcommand{\goal}{\mathbf{g}_t}
\newcommand{\proprio}{\mathbf{s}^\text{prop}_t}
\newcommand{\baserew}{r_t^\text{base}}
\newcommand{\objrew}{r_t^\text{obj}}
\newcommand{\gspace}{\mathcal{G}}
\newcommand{\sspace}{\mathcal{S}}
\newcommand{\aspace}{\mathcal{A}}
\newcommand{\method}{\textbf{ReST-RL}\xspace}
\newcommand{\task}{\textbf{SteadyTray}\xspace}

\title{\LARGE \bf
  SteadyTray: Learning Object Balancing Tasks in Humanoid \\Tray Transport via Residual Reinforcement Learning
}

\author{
  Anlun Huang$^*$, Zhenyu Wu$^*$, Soofiyan Atar$^{\dagger}$, Yuheng Zhi$^{\dagger}$, Michael Yip\\
  UC San Diego \hspace{2em} ${}^* {}^\dagger$ {Equal Contribution}\\ 
  Website: \url{https://steadytray.github.io}
}

\begin{document}

\twocolumn[{
\renewcommand\twocolumn[1][]{#1}%
\maketitle
\vspace{-2em}
\begin{center} 
    \centering
    \includegraphics[width=\textwidth]{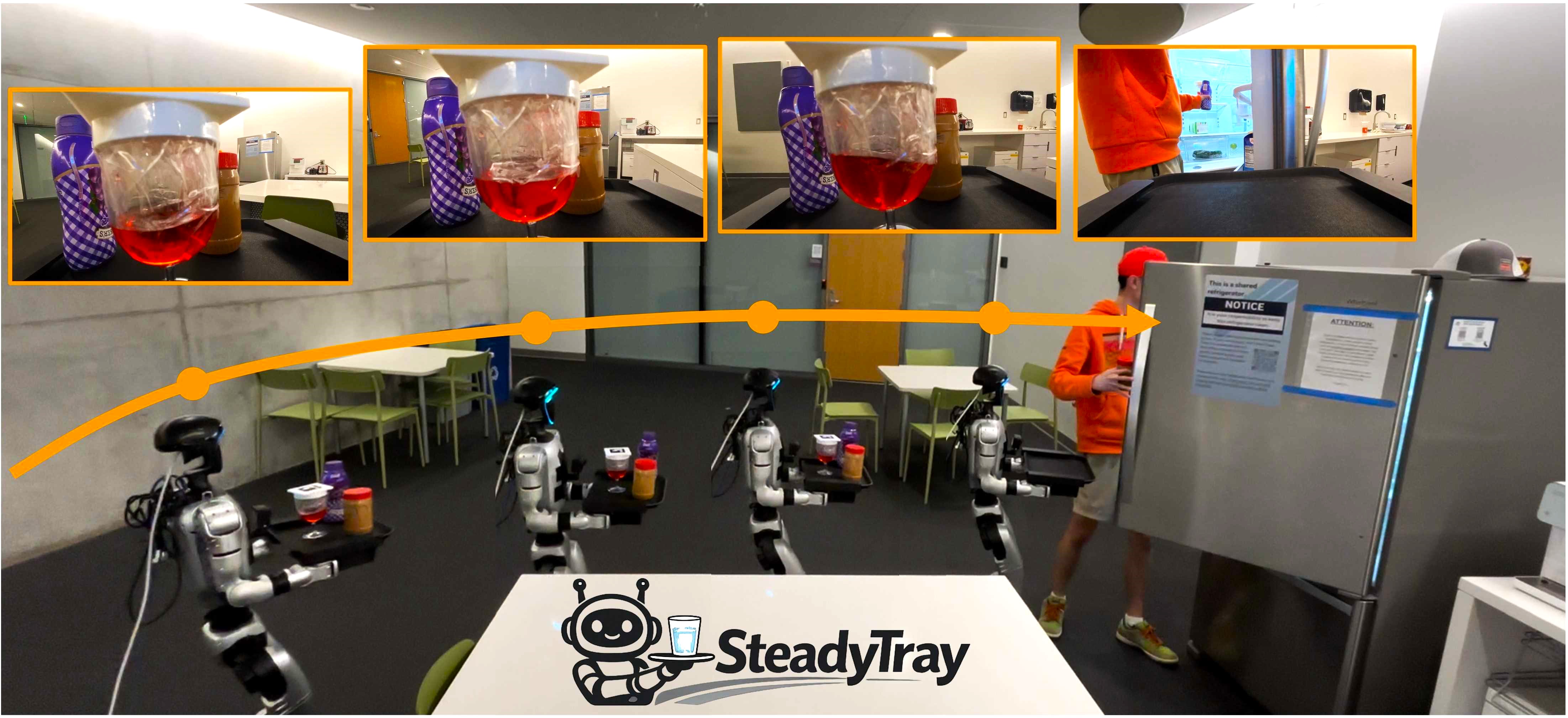}
    \captionof{figure}{\method enables a Unitree G1 humanoid to perform the \task task in a real-world setting, with a fluid-filled wine glass as one of the payload.
    The robot keeps the tray level to prevent fluid sloshing, glass tipping, and payload falling during transport.}
    \label{fig:hero_image}
\end{center}%
}]

\thispagestyle{empty}
\pagestyle{empty}



\begin{abstract}

Stabilizing unsecured payloads against the inherent oscillations of dynamic bipedal locomotion remains a critical engineering bottleneck for humanoids in unstructured environments. To solve this, we introduce \method, a hierarchical reinforcement learning architecture that explicitly decouples locomotion from payload stabilization, evaluated via the \task benchmark. 
Rather than relying on monolithic end-to-end learning, our framework integrates a robust base locomotion policy with a dynamic residual module engineered to actively cancel gait-induced perturbations at the end-effector. 
This architectural separation ensures steady tray transport without degrading the underlying bipedal stability. In simulation, the residual design significantly outperforms end-to-end baselines in gait smoothness and orientation accuracy, achieving a 96.9\% success rate in variable velocity tracking and 74.5\% robustness against external force disturbances. Successfully deployed on the Unitree G1 humanoid hardware, this modular approach demonstrates highly reliable zero-shot sim-to-real generalization across various objects and external force disturbances. 


\end{abstract}

\section{INTRODUCTION}

In December 2025, U.S. employers reported 1.25 million job openings in health care and social assistance and 0.81 million in accommodation and food services
\cite{bls_jolts_dec2025}. The need is pressing to ease labor shortages by automating routine delivery and service tasks \cite{gu2025humanoid}. Practical service tasks such as spill-free meal delivery, sterile instrument transport in operating rooms, and assistance in elder-care facilities require robots to move and carry objects with human‑level reliability in dynamic environments.
These human-centered environments are cluttered and designed for humans, often with equipment, cables, and narrow passages that constrain wheeled or quadrupedal platforms. 
Wheeled platforms cannot generate \textit{active} stabilizing maneuvers when experiencing large external disturbances.
With the rapid progress in humanoid locomotion, humanoid robots may arguably be better suited to navigate such spaces considering their small planar footprint and active stabilization capabilities. 

To maximize transport efficiency and facilitate seamless post-delivery interaction, these scenarios require humanoid robots to walk while balancing unsecured, unstable payloads on trays. 
It remains unsolved despite substantial recent progress in bipedal locomotion \cite{wu2026perceptive} and loco-manipulation \cite{qiu2025humanoid}.
Foot impacts during locomotion are a primary source of torso and base oscillations, and these disturbances propagate through the robot's kinematic chain. Thus, both the upper body and the gait must adapt to cancel them while maintaining balance. 
The most relevant approaches are end-effector stabilization methods such as SoFTA \cite{li2025hold}, which can reduce oscillation of one object hung or attached to the robot's hand during walking to some extent. However, to the best of our knowledge, no methods have demonstrated success in balancing unsecured objects on a tray, such as liquid-filled glasses or fragile instruments, during common maneuvers such as turning, accelerating, decelerating, or withstanding external pushes.

To address this challenge, we formulate \textit{transporting unsecured objects on a tray with a bipedal humanoid robot} (\textbf{\task}) as a bimanual loco-manipulation problem representative of the workforce scenarios above. The key technical difficulty is that the tray introduces a nonrigid intermediate coupling between the robot and payload: the end-effectors regulate tray motion, while objects are only passively constrained and therefore susceptible to slip, tilt, and toppling under dynamic maneuvers. This coupling induces an objective conflict between locomotion and stabilization, since the controller must generate agile whole-body gaits while maintaining near-level end-effector orientation to stabilize tray and payload \cite{li2025hold}. 
We address this with \textbf{Residual Student-Teacher Reinforcement Learning (\method)}, which augments a pre-trained locomotion base policy with a residual module comprising (i) an encoder over privileged robot and payload related observations and (ii) an adapter that outputs corrective residual actions. By freezing the base policy and optimizing only the residual module, \method explicitly separates the two objectives, preserving nominal gait performance while concentrating learning capacity on payload stabilization. 
Then through policy distillation, the encoder is converted from taking privileged observations to taking deployable ones, while the adapter remains fixed. 


This paper provides the following technical contributions:
\begin{itemize}
  \item We propose \method for \task, a residual student--teacher RL framework that augments a pretrained locomotion policy with a payload-stabilization residual adapter for steady tray transport during dynamic walking.
  \item We identify key training design choices, including adding observation delay, control latency, and other domain randomizations, that are critical for robustness to disturbance and sim-to-real transfer.
  \item We evaluate the approach in simulation and real-world deployment on a Unitree G1 humanoid, demonstrating robust transport of diverse unsecured payloads under external disturbances.
  
\end{itemize}

\section{Related Work}

\subsection{Humanoid Loco-Manipulation}
Loco-manipulation tasks require humanoids to achieve locomotion and manipulation objectives simultaneously, coupling gait stability with contact-rich object interaction under shared dynamical constraints \cite{gu2025humanoid}. Recent sim-to-real reinforcement learning approaches have demonstrated autonomous pick and place\cite{he2025viral}, door opening\cite{xue2025opening}, package carrying and placement\cite{zhao2025resmimic,weng2025hdmi,fu2025demohlm}, and collaborative transport\cite{du2025learning}. 

Prior works can be viewed through how they inject supervision to reduce the exploration burden of contact-rich whole-body learning, including human-guided imitation (e.g., teleoperation, mocap, or human videos) \cite{zhao2025resmimic,yin2025visualmimic}, interaction-preserving retargeting or trajectory optimization \cite{yang2025omniretarget,liu2024opt2skill}, and privileged teacher--student distillation \cite{he2025viral,xue2025opening}. 
While these methods improve robustness and generalization, stable transport of \textit{unsecured} payloads during walking remains underexplored due to non-stationary contact and disturbance-sensitive object dynamics.

\subsection{Humanoid Policy Architectures}
Humanoid whole-body control (WBC) policies are typically implemented either as monolithic end-to-end RL controllers \cite{he2024learning,ji2024exbody2} or as structured multi-agent designs that decouple locomotion and upper-body control \cite{zhang2025falcon,lu2025mobile}. Training commonly relies on teacher–student distillation and PPO-style actor–critic optimization \cite{hinton2015distilling,schulman2017ppo}.

Residual reinforcement learning augments pretrained locomotion policies with corrective modules to improve robustness under modeling errors and disturbances. Prior works apply residual adaptation for sim-to-real transfer\cite{he2025asap}, robust teleoperation and motion tracking\cite{sun2026mosaic,zhang2025track}, precise interaction and collaborative carrying \cite{zhao2025resmimic,du2025learning}. In contrast, we adopt residual RL on stabilization of \task, where object-level dynamics must be regulated without degrading gait stability.

\subsection{Stability Control in Mobile Manipulation}
Stability control during transport has been studied for wheeled and aerial manipulators using model-based coordination \cite{pankert2020perceptive,he2025flying}. On legged systems, quadrupedal payload transport has been explored using tactile feedback \cite{lin2025locotouch}. For humanoids, SoFTA \cite{li2025hold} demonstrates end-effector stabilization during walking. However, steady transport of \textit{unsecured} objects requires coordinated whole-body adaptation beyond upper-body compensation alone. Our work addresses this gap through residual whole-body stabilization.
                                                                                          
\section{METHODOLOGY}

\begin{figure*}[thpb]
  \centering
  \includegraphics[width=0.98\textwidth]{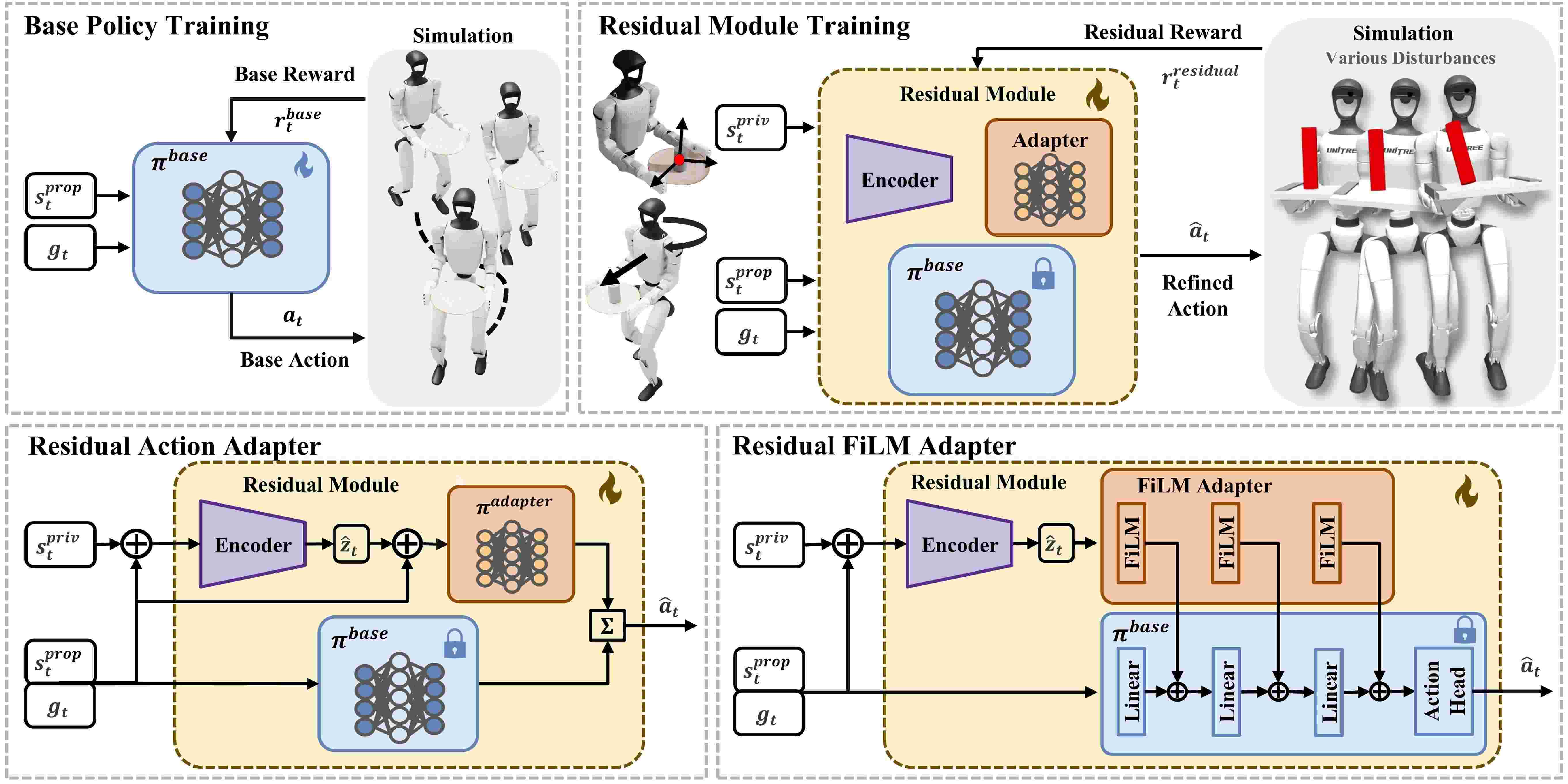}
\caption{Overview of the \textbf{ReST-RL} framework. \textbf{Base Policy Training:} A locomotion policy is first trained to carry a tray while maintaining a stable gait. \textbf{Residual Module Training:} using privileged observations, a residual module learns whole-body corrective adjustments on top of the frozen base policy to stabilize the payload under disturbances. Two residual designs are considered: (a) \textbf{Residual Action Adapter}, which adds corrective residual actions to the base action, and (b) \textbf{Residual FiLM Adapter}, which modulates intermediate activations of the frozen base policy via layer-wise FiLM residuals. The student encoder distillation process is shown in Fig.~3.}

  \label{fig:method}
\end{figure*}

\subsection{Problem Statement}

This work aims to achieve stable dual-arm load transport during humanoid robot locomotion by maintaining the tray level and resisting object tipping under disturbances. 
The problem is formulated as a goal-conditioned reinforcement learning problem with policy $\pi: \gspace \times \sspace \to \aspace$, where $\gspace$, $\sspace$, and $\aspace$ denote the goal, observation, and action spaces, respectively. $\aspace \subseteq \mathbb{R}^{n}$ is target joint positions for all humanoid joints.
At time $t$, the goal is denoted $\goal = [\mathbf{\hat{v}}^{r_x}_t, \mathbf{\hat{v}}^{r_y}_t, \boldsymbol{\hat{\omega}}^{r_\text{yaw}}_t] \in \gspace$ which consists of desired robot linear and angular velocity.
The observation is $\mathbf{s}_t = (\proprio, \objobs) \in \sspace$, including proprioceptive data and object data. The proprioceptive data $\proprio = [\mathbf{q}_{t-H:t}, \mathbf{\dot{q}}_{t-H:t}, \boldsymbol{\omega}^\text{r}_{t-H:t}, \mathbf{g}^\text{r}_{t-H:t}, \mathbf{a}_{t-H:t-1}] \in \mathcal{S}^P$ contains $H$-step histories of joint positions $\mathbf{q}_t \in \mathbb{R}^n$, joint velocities $\mathbf{\dot{q}}_t \in \mathbb{R}^n$, robot angular velocity $\boldsymbol{\omega}^\text{r}_t \in \mathbb{R}^3$, robot projected gravity $\mathbf{g}^\text{r}_t \in \mathbb{R}^3$, and action $\mathbf{a}_t \in \mathbb{R}^n$. The object data $ \objobs = [\,\mathbf{p}^{c}_{t-H:t}, \mathbf{q}^{c}_{t-H:t}\,]$ represents the object pose in the camera frame ${c}$ , where $\mathbf{p}^{c}_t = (x^{c}_t, y^{c}_t, z^{c}_t) \in \mathbb{R}^3$ is the position and $\mathbf{q}^{c}_t \in \mathbb{R}^4$ is the quaternion orientation.

\subsection{Base Policy Training}
As shown in Fig.~\ref{fig:method}, \method framework starts by training a base policy $\pi^\text{base}: \gspace \times \sspace^P \to \aspace$ to perform robust locomotion while maintaining a level tray. The base policy takes the $H=5$ steps observation $\proprio$ and outputs the base action $a_t$, optimized with the goal-conditioned objective $r^\text{base}_t = \mathcal{R}^\text{base}(\goal, \proprio)$. The policy is updated via PPO\cite{schulman2017ppo} to maximize the cumulative discounted reward:
$
\mathbb{E} \left[ \sum_{t=1}^{T} \gamma^{t-1} r^\text{base}_t \right]
$.

\subsection{Residual Module Learning}
As shown in Fig.~\ref{fig:method}, after obtaining the base policy $\basepolicy$, we train an encoder--adapter residual module that encodes observations and applies corrective updates to the frozen base policy in $\aspace$ for robust object stabilization.
The residual module is first trained with privileged observations and then distilled to one without the privileged information for real-world deployment. This approach draws inspiration from previous work on rapid motor adaptation with privileged information \cite{qi2023general}.

Privileged observations are defined as $\privobs = [\mathbf{v}^{\text{r}}_{t-H:t}, \mathbf{p}^{\text{tray}}_{t-H:t}, \mathbf{g}^{\text{tray}}_{t-H:t}, \mathbf{p}^{\text{obj}}_{t-H:t}, \mathbf{v}^{\text{obj}}_{t-H:t}, \boldsymbol{\omega}^{\text{obj}}_{t-H:t}, \mathbf{g}^{\text{obj}}_{t-H:t}]$, which contain the robot linear velocity, tray position and projected gravity, and the object position, linear and angular velocities, and projected gravity, all lying in $\mathbb{R}^3$.
We concatenate the proprioceptive history $\proprio$, the goal $\goal$, and the privileged observations $\privobs$ as the encoder input to provide robot-state context and task intent.
A longer temporal window ($H=32$) is used for all observations $(\proprio, \privobs)$ to capture gradual destabilization trends that are not evident from short-horizon observations.
The encoder outputs a 64-dim feature vector $\hat{\mathbf{z}}_t = \phi(\privobs, \proprio, \goal) \in \mathcal{Z},$ where $\mathcal{Z}$ denotes the latent feature space.

As shown in Fig. \ref{fig:method}, two representative integration mechanisms are adopted here to explore residual refinement under different structural configurations. 

\subsubsection{\textbf{Residual Action Adapter}}
The action adapter is defined as $\residualpolicy : \hat{\mathcal{S}} \to \aspace$, where $\hat{\mathcal{S}} := \mathcal{Z} \times \mathcal{S}^P \times \mathcal{G}$. At time $t$, the input is $\hat{\mathbf{s}}_t = [\hat{\mathbf{z}}_t;\proprio;\goal] \in \hat{\mathcal{S}}$, and the action adapter produces a corrective action $\tilde{\mathbf{a}}_t = \residualpolicy(\hat{\mathbf{s}}_t) \in \aspace$, which is applied on top of the base action $\mathbf{a}_t$ from the pretrained base policy. The final refined action sent to the low-level controller is
\begin{equation}
\hat{\mathbf{a}}_t
= \alpha_\text{base}\,\mathbf{a}_t
+ \alpha_\text{residual}\,\tilde{\mathbf{a}}_t
+ \mathbf{q}_\text{default}.
\label{eq:residual_action}
\end{equation}
where $\alpha_\text{base}$ is the action scale for base action, $\alpha_\text{residual}$ is the action scale for residual action, and $q_\text{default}$ is the default robot joint position. 

\subsubsection{\textbf{Residual FiLM Adapter}}
We use FiLM adapters \cite{film} to modulate frozen layers of the base policy through affine, feature-wise conditioning from the latent feature $\hat{\mathbf{z}}_t$. For a frozen linear layer $f(\cdot)$ in $\basepolicy$ with output $\mathbf{y}=f(\mathbf{x})$, an adapter predicts $(\boldsymbol{\gamma}_t,\boldsymbol{\beta}_t)=\mathrm{FiLM}(\hat{\mathbf{z}}_t)$ and computes
\begin{equation}
y'_i = y_i \bigl(1+\gamma_{t,i}\bigr) + \beta_{t,i},\quad i=1,\dots,d,
\label{eq:film}
\end{equation}
where $d$ is the output dimensionality of the modulated layer. Each modulated base policy layer is paired with one FiLM adapter.

In both instantiations of the residual module, the adapter is initialized to output zero correction at the start of training, so the initial behavior matches the pre-trained base policy and preserves its locomotion skill. As training progresses, the residual modules gradually introduce task-specific refinements.
PPO jointly optimizes adapter parameters $\theta_\text{adapter}$ and encoder parameters $\theta_\text{encoder}$ to maximize
$\mathbb{E}\!\left[\sum_{t=1}^{T} \gamma^{t-1} r_t^\text{residual}\right]$.
The reward $r_t^\text{residual}$ is defined in Section~\ref{section:reward}.

\subsection{Residual Module Distillation}
The privileged information taken by the encoder of the residual module is not directly observable in the real world.
Therefore, a non-privileged state encoder is distilled from the privileged state encoder using DAgger\cite{2011dagger} for real-world deployment.

\begin{figure}[t]
  \centering
  \includegraphics[width=0.99\linewidth]{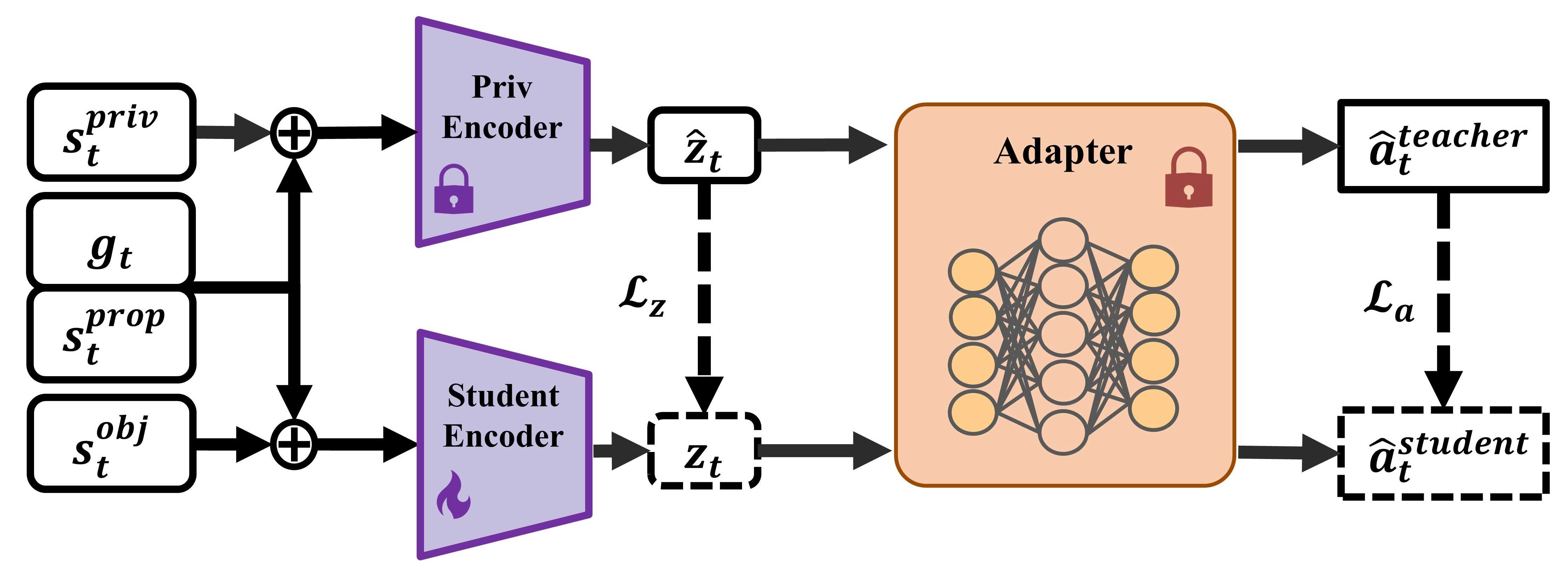}

  \caption{\textbf{Residual module distillation.} The teacher encoder uses privileged observations, whereas the student encoder uses object-centric inputs; both feed into a frozen residual adapter for latent alignment.}
  \label{fig:distill}
\end{figure}

As shown in Fig \ref{fig:distill}, the distillation process focuses exclusively on the encoder, while the adapter remains frozen. 
The student encoder receives only object-centric observations $\objobs$ (with a 32-step history) and outputs $\mathbf{z}_t = \phi'(\objobs, \proprio, \goal)$. The frozen adapter maps $\mathbf{z}_t$ to refined student actions $\hat{\mathbf{a}}^\text{student}_t$. These trajectories are collected and stored during rollout. The teacher encoder, trained with privileged observations $\privobs$, outputs latent features $\hat{\mathbf{z}}_t = \phi(\privobs, \proprio, \goal)$, which are passed through the same frozen adapter to produce refined teacher actions $\hat{\mathbf{a}}^\text{teacher}_t$. 
The student encoder is then trained with a joint loss that aligns both latent features and refined actions:
\begin{equation}
  \mathcal{L} = \mathcal{L}_z + \mathcal{L}_a = \| \mathbf{z}_t - \hat{\mathbf{z}}_t \|_2^2 + \| \hat{\mathbf{a}}^\text{student}_t - \hat{\mathbf{a}}^\text{teacher}_t \|_2^2,
  \label{eq:distill_loss}
\end{equation}

\subsection{Domain Randomization and Environment}


Robot, tray, and object physical parameters are randomized during training, including mass, friction, restitution, and the robot torso center of mass. This exposes the policy to diverse contact dynamics and inertial variations. Control delay is also introduced to account for actuation and communication latency.

Cylindrical objects with diverse scales are sampled during training and initialized near the tray center with randomized horizontal offsets and yaw rotation to avoid overfitting to a fixed initial configuration. To ensure a stable initialization phase, velocity commands are set to zero within the initial second of each episode \cite{lin2025locotouch}. Random push perturbations are applied to the object during training to improve robustness and encourage the residual module to learn recovery behaviors under external disturbances.


Observation noise is injected into all observation terms during both training and distillation. To model perception latency, an observation delay is introduced for object-related observations $(\objobs, \privobs)$. At time $t$, a delay $\ell_t$ is sampled such that the residual module receives measurements from $t-\ell_t$, and the sampled delay is maintained for multiple steps to mimic temporally correlated latency. This delay is applied only to object observations, as object perception typically incurs higher latency than proprioceptive sensing.


\subsection{Reward Design}
\label{section:reward}
\begin{table}[t]

  \caption{Key Rewards Functions}
  \label{tab:reward_tb}
  \centering
  \begin{tabular}{ll}
    \toprule
    Term & Equation \\
    \midrule
    Object Upright & $\exp(-\lambda_{\text{upright}} \big\|\mathbf{P}_{xy}\big(\mathbf{g}^\text{obj}_t)\big\|^2_{2})$ \\
    Object-tray Contact & $\mathbf{1}_{\{\text{object\_tray\_contact}\}}$ \\
    \midrule
    Linear XY Vel Tracking & $\exp(-\lambda_{v_{xy}} \big\|\mathbf{v}^{r_{x,y}}_t -\mathbf{\hat{v}}^{r_{x,y}}_t\big\|^2_{2})$  \\
    Angular Vel Tracking & $\exp(-\lambda_{\boldsymbol{\omega}_z} \big\|\boldsymbol{\omega}^{r_\text{yaw}}_t -\boldsymbol{\hat{\omega}}^{r_\text{yaw}}_t\big\|^2_{2})$  \\
    Torso Vertical Vel & $\exp(-\lambda_v v^{r_z\ 2}_t)$ \\
    Torso Angular Vel & $\exp(-\lambda_\omega \big\|\boldsymbol{\omega}^{r_{x,y}}_t \big\|^2_{2})$ \\
    Feet Impact & $\exp(-\lambda_{v_z}\left\|\mathbf{v}^{\text{feet}}_{z,t}-\mathbf{v}^{\text{feet}}_{z,t-1}\right\|_2^2)$\\
    Tray Upright & $\exp(-\lambda_{\text{upright}} \big\|\mathbf{P}_{xy}\big(\mathbf{g}^\text{tray}_t)\big\|^2_{2})$ \\
    Tray-EE Contact & $\mathbf{1}_{\{\text{tray\_ee\_contact}\}}$ \\
    Tray-EE Quat Error & $\exp(-\lambda_{\text{quat}}\Delta\theta_t^2)$ \\
    \bottomrule
  \end{tabular}
\end{table}

Several base-locomotion reward terms are adapted from prior open-source works \cite{unitree_rl_lab,2025olaf}, and augmented with object-stabilization terms.
We define a base reward $\baserew$ and a stabilization reward $\objrew$.
The base policy is optimized with $\baserew$, while the residual module is optimized with the combined objective $r_t^\text{residual}=\baserew+\objrew$, where $\objrew$ promotes payload stability during locomotion.
TABLE~\ref{tab:reward_tb} summarizes the key reward components used in \method. 


\begin{table*}[t]
  \centering
  \small
  \footnotesize
  \setlength{\tabcolsep}{4pt}
  \begin{tabular}{llccccccc}
    \toprule
    \multicolumn{2}{c}{\textbf{Simulation Results (Isaac Lab)}} &
    \multicolumn{2}{c}{\textbf{TrackLinErr (m/s)} $\mathbf{\downarrow}$} &
    \multicolumn{2}{c}{\textbf{TrackAngErr (rad/s)} $\mathbf{\downarrow}$} &
    \multicolumn{2}{c}{\textbf{Grav-XY} $\mathbf{\downarrow}$} &
    \multirow{2}{*}{\textbf{Success Rate (\%)} $\mathbf{\uparrow}$} \\
    \cmidrule(lr){3-4}\cmidrule(lr){5-6}\cmidrule(lr){7-8}
    \textbf{Task} & \textbf{Method} &
    \textbf{mean} & \textbf{std} &
    \textbf{mean} & \textbf{std} &
    \textbf{mean} & \textbf{std} &
    \textbf{} \\
    \midrule
    \multirow{5}{*}{Command Track}
    & Base Policy (WB)        & 0.110 & 0.107 & \textbf{0.078} & 0.065 & 0.179 & 0.064 & 47.4 \\
    & End2End            & 0.116 & 0.111 & 0.096 & 0.078 & 0.046 & 0.029 & 89.1 \\
    & \method (Action WB) & \textbf{0.093} & 0.100 & 0.081 & 0.070 & \textbf{0.029} & 0.022 & 95.9 \\
    & \method  (Action JT) & 0.160 & 0.102 & 0.135 & 0.106 & 0.031 & 0.019 & 96.7 \\
    & \method \textbf{(FiLM WB)}     & 0.106 & 0.098 & 0.069 & 0.058 & 0.046 & 0.023 & \textbf{96.9} \\
    \midrule
    \multirow{5}{*}{Push Robot}
    & Base Policy (WB)        & 0.162 & 0.222 & 0.110 & 0.156 & 0.190 & 0.089 & 9.1 \\
    & End2End            & 0.170 & 0.215 & 0.123 & 0.155 & 0.055 & 0.051 & 44.0 \\
    & \method (Action WB) & 0.146 & 0.204 & 0.110 & 0.142 & 0.039 & 0.044 & 73.4 \\
    & \method (Action JT)  & 0.190  & 0.168 & 0.176 & 0.213 & \textbf{0.038} & 0.040 & 72.5 \\
    & \method \textbf{(FiLM WB)}     & \textbf{0.142} & 0.168 & \textbf{0.094} & 0.123 & 0.043 & 0.032 & \textbf{84.6} \\
    \midrule
    \multirow{5}{*}{Push Object}
    & Base Policy (WB)        & 0.110 & 0.111 & 0.079 & 0.066 & 0.186 & 0.072 & 25.2  \\
    & End2End            & 0.117 & 0.114 & 0.099 & 0.080 & 0.049 & 0.043 & 50.2 \\
    & \method (Action WB) & \textbf{0.096} & 0.103 & 0.084 & 0.076 & \textbf{0.023} & 0.039 & 71.3 \\
    & \method (Action JT)  & 0.161 & 0.104 & 0.141 & 0.118 & 0.033 & 0.039 & 70.7 \\
    & \method \textbf{(FiLM WB)}     & 0.107 & 0.100 & \textbf{0.073} & 0.068 & 0.040 & 0.034 & \textbf{74.6} \\
    \bottomrule
  \end{tabular}
  \caption{Simulation results in Isaac Lab. Object stability of baselines and instantiations of \method with different Residual Module architectures is evaluated across various scenarios. In all three tasks, the target locomotion commands are identical.}
  \label{tab:sim-results}
\end{table*}

The base reward $\baserew$ primarily consists of standard locomotion objectives, with regularization terms (e.g., joint limits and action rate penalties) to prevent aggressive motions.
Torso stabilization terms reward low vertical torso velocity and low roll/pitch angular velocity, promoting smooth upper-body motion during locomotion.
Waist-joint deviation is only lightly penalized, since overly strict waist constraints reduce the policy's ability to compensate for disturbances through active upper-body adjustment.
A foot impact term rewards small changes in vertical foot velocity, reducing landing forces propagated to the tray.
The base reward also includes tray-related terms.
In particular, a tray--end-effector orientation term rewards maintaining a predefined holding configuration, ensuring consistent contact between the tray and both end-effectors.

For the object reward $\objrew$, we use a sparse formulation which includes two object-related terms: an upright term that encourages alignment between the object's vertical axis and gravity, and a contact term that rewards object-tray contact. Without explicit position or velocity targets, this design allows adapter to discover emergent recovery strategies under perturbations.


\section{EXPERIMENTS}

This section assesses \method's performance through simulation studies and real-world deployment on a Unitree G1 humanoid robot (29 DoF). In particular, evaluation focus on answering following questions:

\textbf{Q1:} How do key components (action adapter, FiLM adapter, and observation delay) affect performance on \task task?

\textbf{Q2:} How robust is \method to disturbances of different types and magnitudes, and how well does it generalize across objects with diverse scales?

\textbf{Q3:} Does \method demonstrate reliable sim-to-real transfer?




\subsection{Experiment Setup}
All experiments are conducted on Unitree G1 robots with specially designed end effectors for tray holding.
The end-effectors support the tray and partially constrain its motion, but the tray is not rigidly attached to either end-effector.
In real-world deployment, we place an AprilTag on the object and use the Unitree G1's head-mounted RealSense D435 camera to estimate the object's pose. 

\subsubsection{\textbf{Tasks}}
We evaluate three simulation tasks: (i) Command Track, which tracks velocity commands while carrying a payload, with targets resampled every 10s; (ii) Push Robot, which applies random forces to the robot torso during locomotion for $t_r$ seconds per event; and (iii) Push Object, which applies random forces to the object during locomotion for $t_o$ seconds per event. 
All three tasks share the same goal-tracking locomotion commands. 
Each episode lasts 20s, and pushes are triggered every 5s. Physical properties and randomization settings match training, and object body scales are randomized.

\subsubsection{\textbf{Baselines}}
Two pre-training base policies are considered: Joint-Training (JT) and Whole-Body (WB). Both follow a two-stage curriculum: (i) a locomotion stage trains a stable gait with observation $(\proprio, \goal)$, then (ii) tray-stability rewards are added while continuing optimization. JT decouples the action space into an upper-body policy (arm/wrist joints) and a lower-body policy (leg joints), while WB uses a single policy controlling all joints. We also consider an End-to-End (End2End) baseline, where a single policy is trained from scratch on both locomotion and stabilization rewards, taking the full observation $(\proprio, \objobs, \goal)$ as input.

\begin{figure}[t]
  \centering
  \begin{subfigure}[t]{0.48\linewidth}
    \centering
    \includegraphics[width=\linewidth]{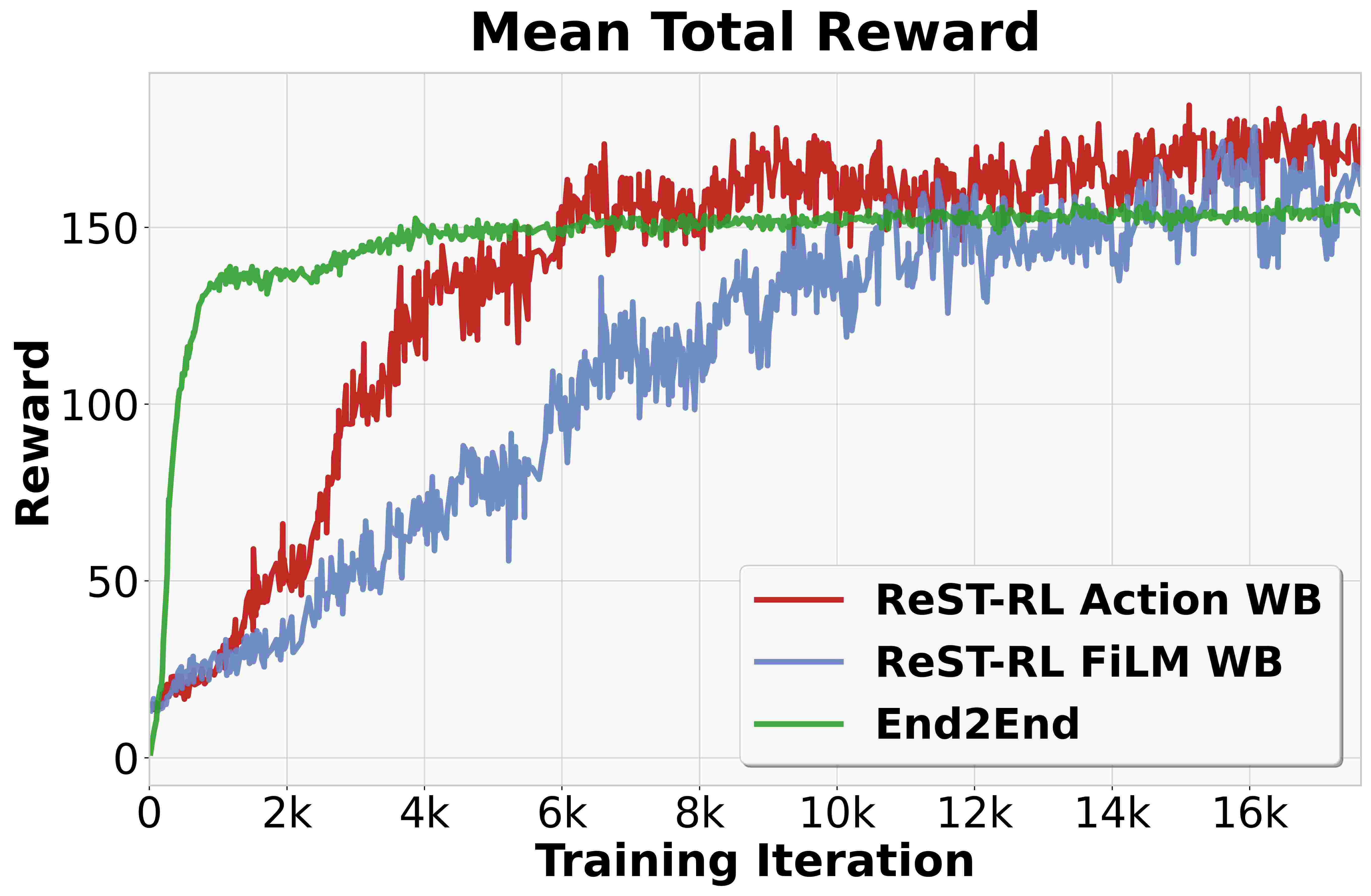}
    \caption{Mean total reward}
    \label{fig:mean_reward}
  \end{subfigure}
  \hfill
  \begin{subfigure}[t]{0.48\linewidth}
    \centering
    \includegraphics[width=\linewidth]{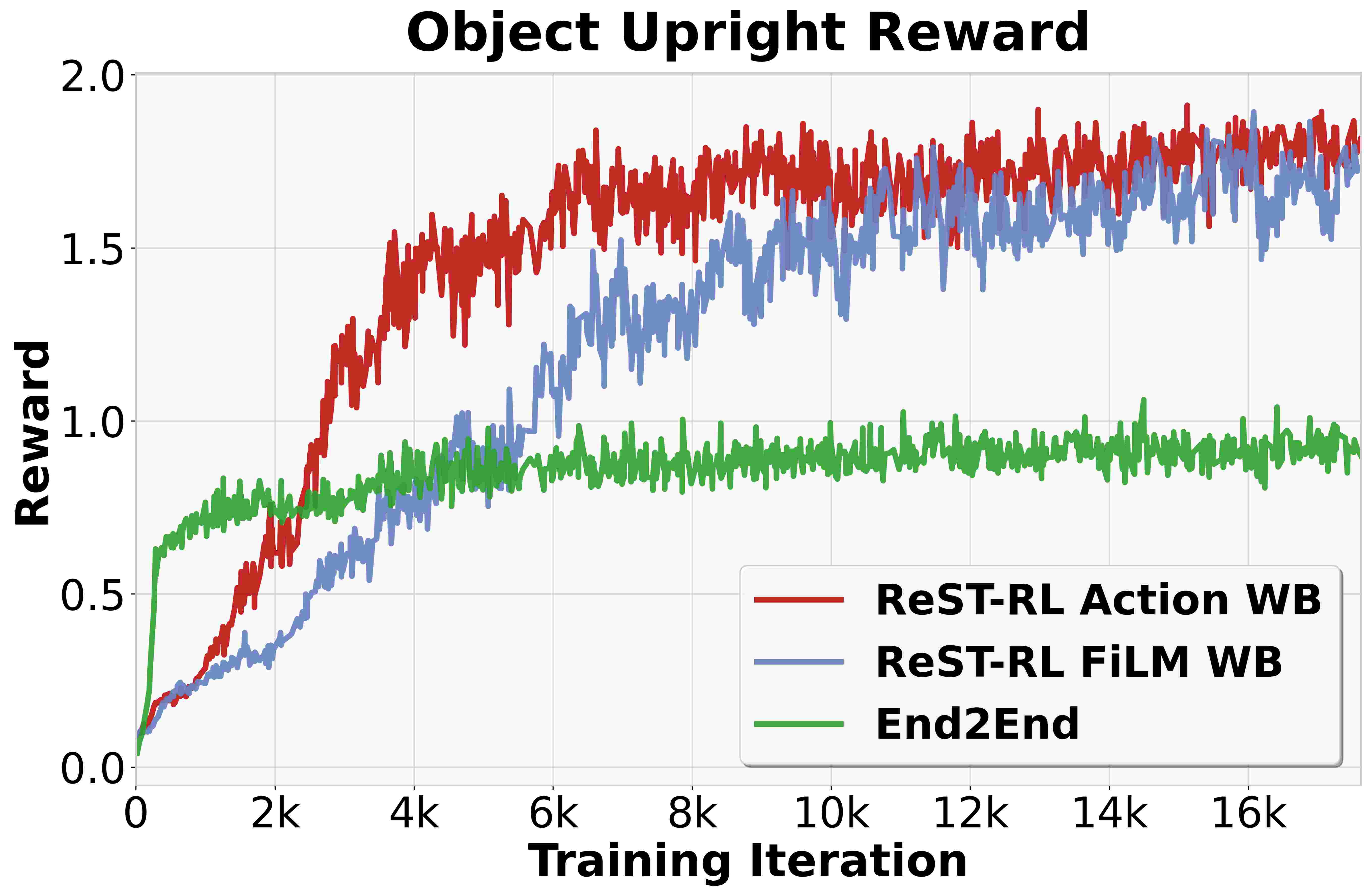}
    \caption{Object upright reward}
    \label{fig:upright_reward}
  \end{subfigure}
  \caption{Training reward comparison between End2End and \method.}
  \label{fig:reward_curves}
\end{figure}

\subsubsection{\textbf{Metrics}}
Success rate is the fraction of trials in which the on-tray object remains upright and stays on the tray for the entire horizon (i.e., no tilt or drop).
Linear and angular tracking errors are the absolute differences between the commanded and measured torso linear speed in the horizontal plane and torso yaw rate about the vertical axis, respectively.
Object Projected Gravity XY Tilt measures object tilt as the $L_2$ norm of the $x,y$ component of the unit gravity vector projected into the object's local frame.
A value of 0 indicates a perfectly upright object, while a value approaching 1 indicates that the object is nearly horizontal.
All simulation metrics are computed over $N$ independent trials.

\subsection{Ablation Study}
TABLE~\ref{tab:sim-results} reports baselines and \method results across three simulation tasks, addressing \textbf{Q1}. 
To isolate design effects, we evaluate: \method (FiLM), which employs a FiLM adapter in the residual module; \method (Action WB) and \method (Action JT), which use action adapters built on WB and JT bases. 

All \method instantiations substantially improve success rates and reduce projected gravity compared to the base policy, while command tracking error remains comparable. This shows the proposed design enhanced object stability without substantially sacrificing command-following performance.
Performance across FiLM WB, Action WB, and Action JT is broadly similar, suggesting that the improvement primarily stems from structured residual adaptation conditioned on object-related features rather than a specific architectural choice. 

\begin{figure}[!t]
  \centering
  \includegraphics[width=0.7\linewidth]{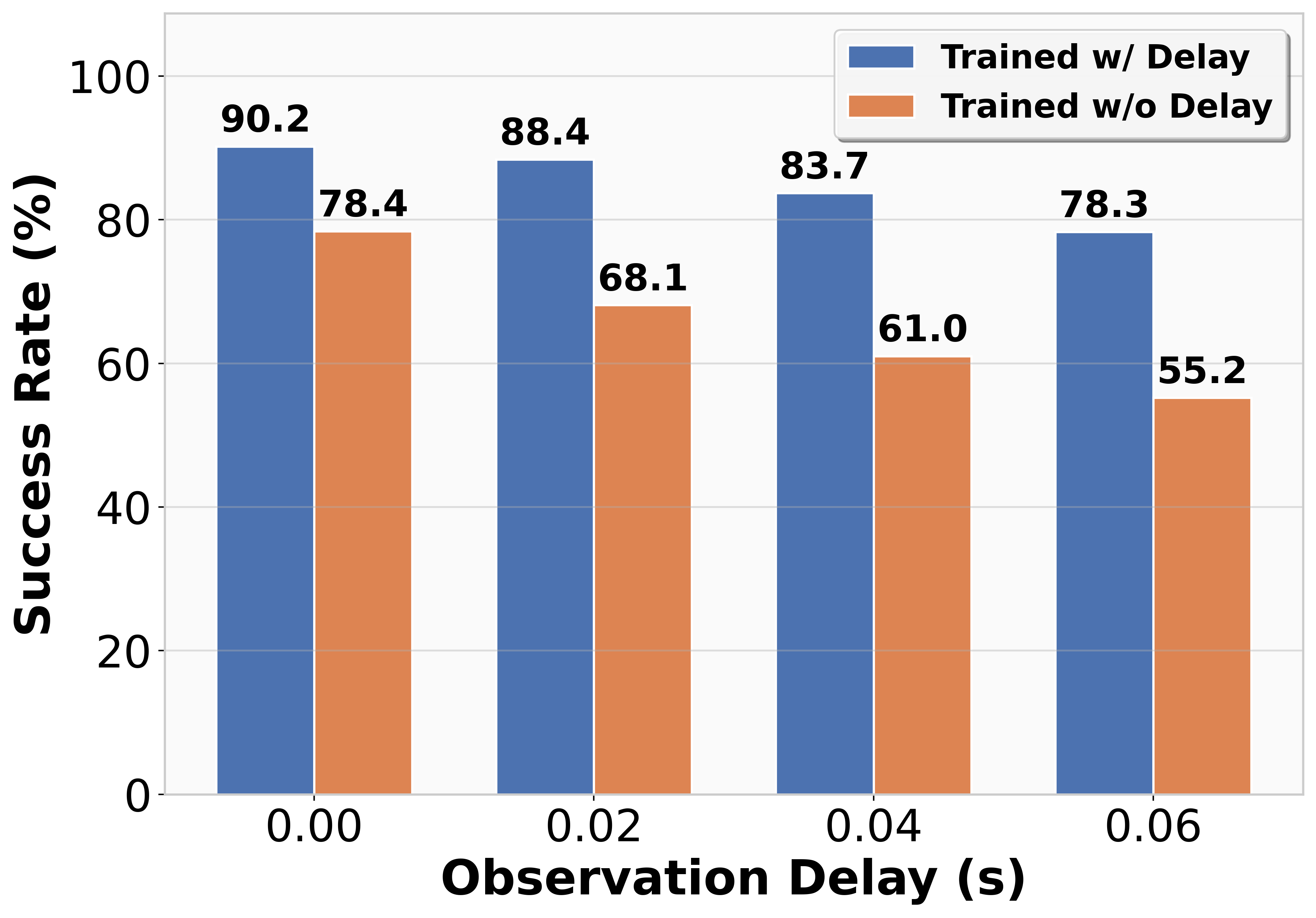}
  \caption{Success rate of \method trained with and without observation delay under increasing perception latency in Push Robot task.}
  \label{fig:delay}
\end{figure}

\begin{figure}[t]
  \centering
  \includegraphics[width=0.95\linewidth]{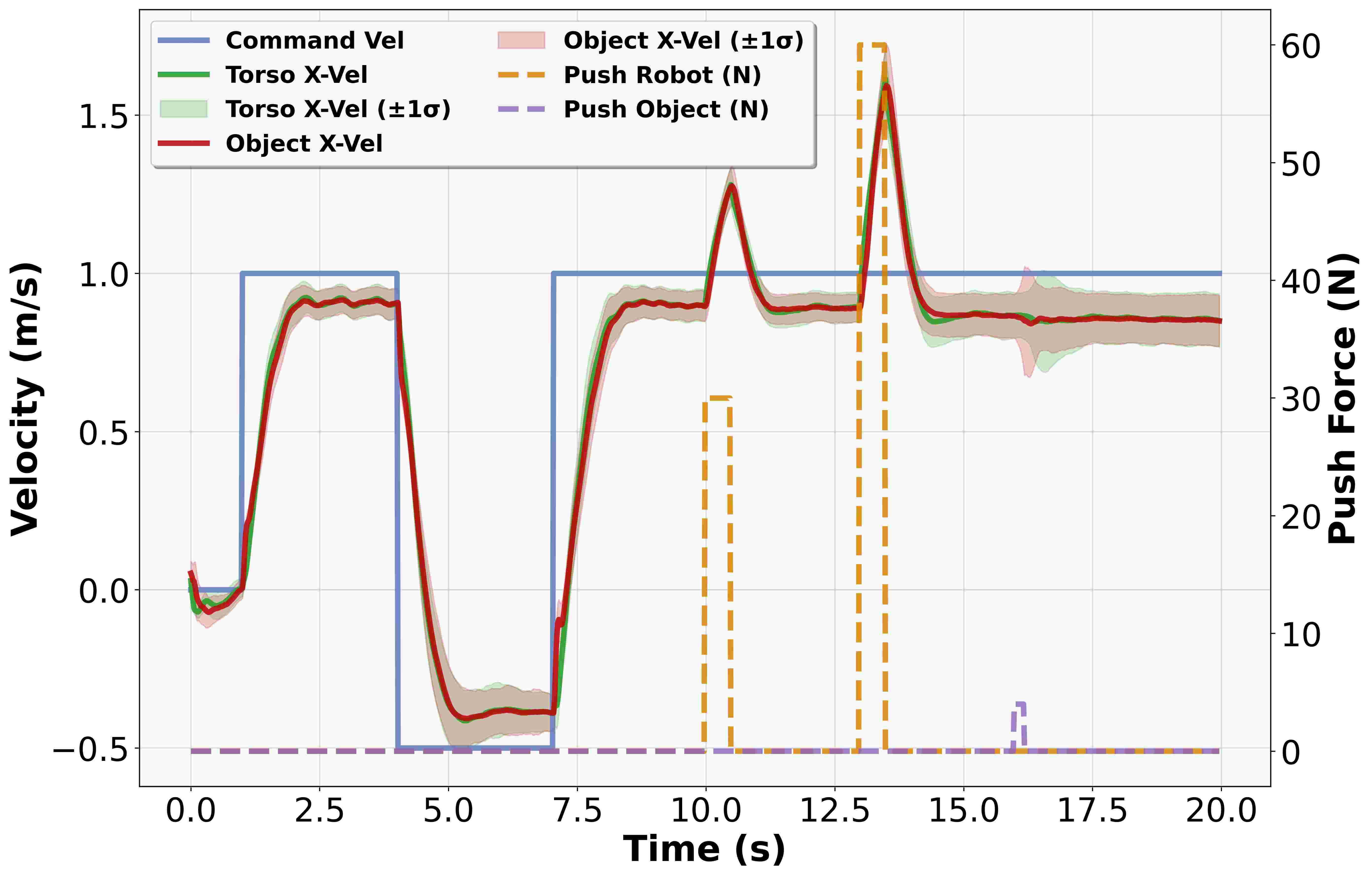}
  \caption{Velocity tracking and disturbance recovery of \method. 
Torso and object velocities closely follow the commanded velocity under command changes and external pushes.}
  \label{fig:obj_robot_vel}
\end{figure}

Although the End2End policy achieves reasonable performance in the Command Track task, its robustness under Push Robot and Push Object is significantly lower than that of \method. 
To better understand this gap, Fig.~\ref {fig:reward_curves} decomposes the reward components during training. 
While End2End attains a comparable mean total reward, much of this reward stems from locomotion-related terms. 
In contrast, its object upright reward remains noticeably lower than that of \method. 
This gap suggests that directly optimizing both locomotion and object-stability objectives is non-trivial, whereas structured residual adaptation provides a more reliable route to robust stabilization.

In addition to structured residual adaptation, we further analyze the effect of delayed observation. 
Fig.~\ref{fig:delay} evaluates the success rate of \method trained with and without observation delay under increasing perception latency. 
Across all delay settings, the \method trained with delayed observations consistently achieves higher success rates, including the zero-latency case. 
Although observation delay is primarily introduced to improve sim-to-real transfer by accounting for sensing latency, these results indicate that it also enhances overall stabilization performance.

\begin{figure}[t]
  \centering
  \begin{subfigure}[t]{0.43\linewidth}
    \centering
    \includegraphics[width=\linewidth]{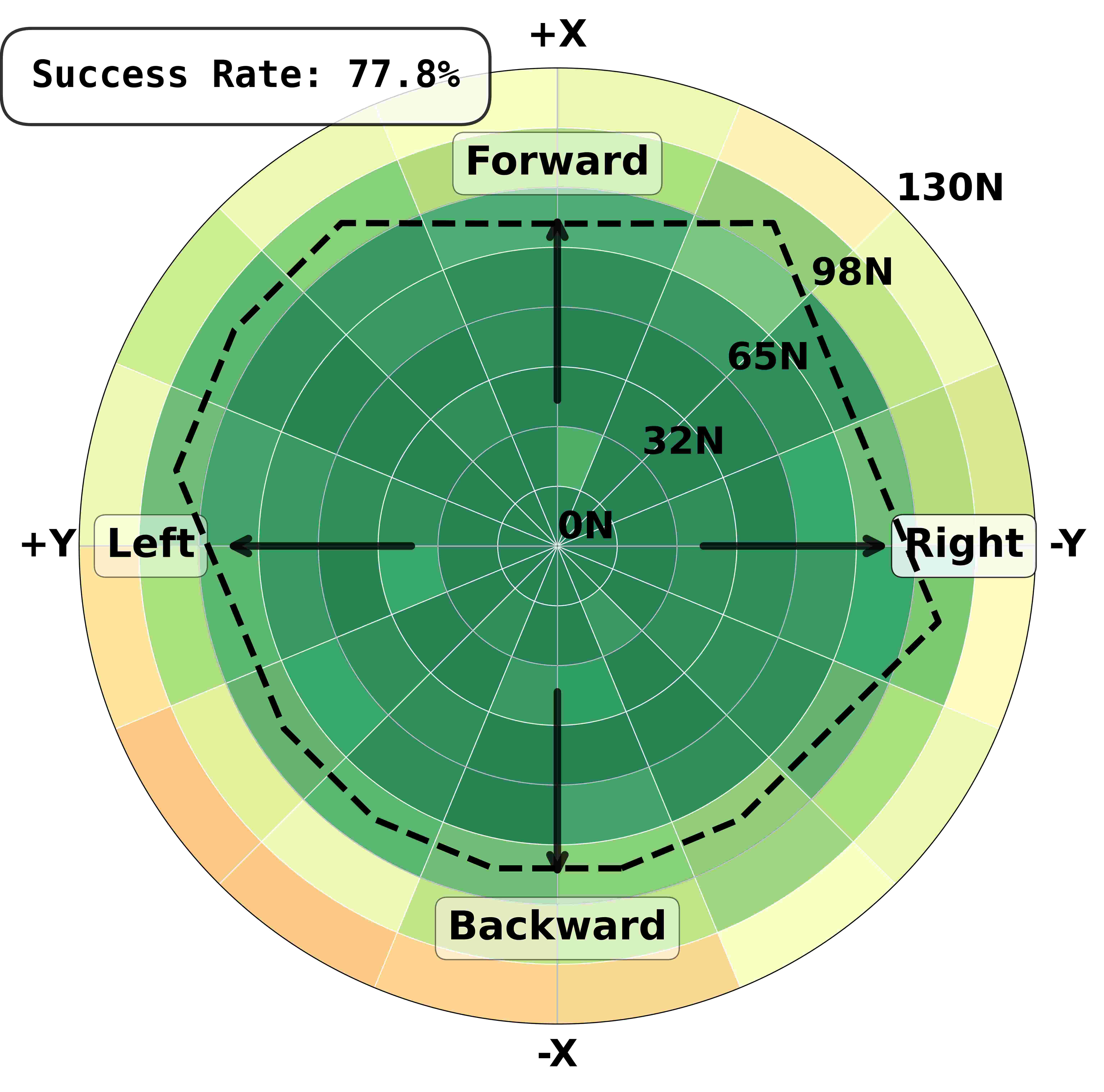}
    \caption{Push Robot}
    \label{fig:push_robot_polar}
  \end{subfigure}
  \begin{subfigure}[t]{0.55\linewidth}
    \centering
    \includegraphics[width=\linewidth]{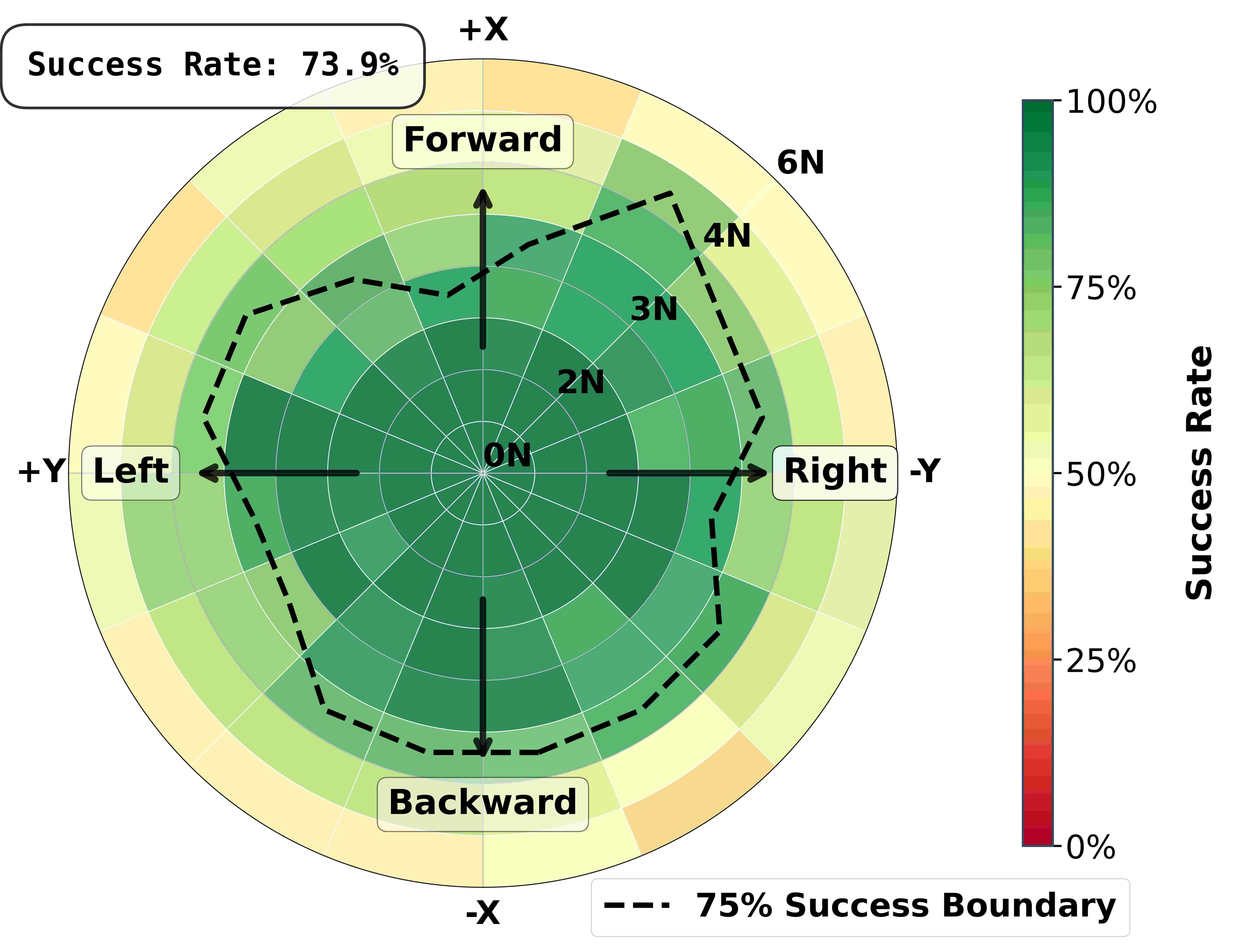}
    \caption{Push Object}
    \label{fig:push_object_polar}
  \end{subfigure}
  \caption{Success rate of \method in (\textbf{a}) Push Robot task and (\textbf{b}) Push Object task under varying push directions and magnitudes.  }
  \label{fig:push_polar}
\end{figure}
\begin{figure}[t]
  \centering
  \includegraphics[width=0.9\linewidth]{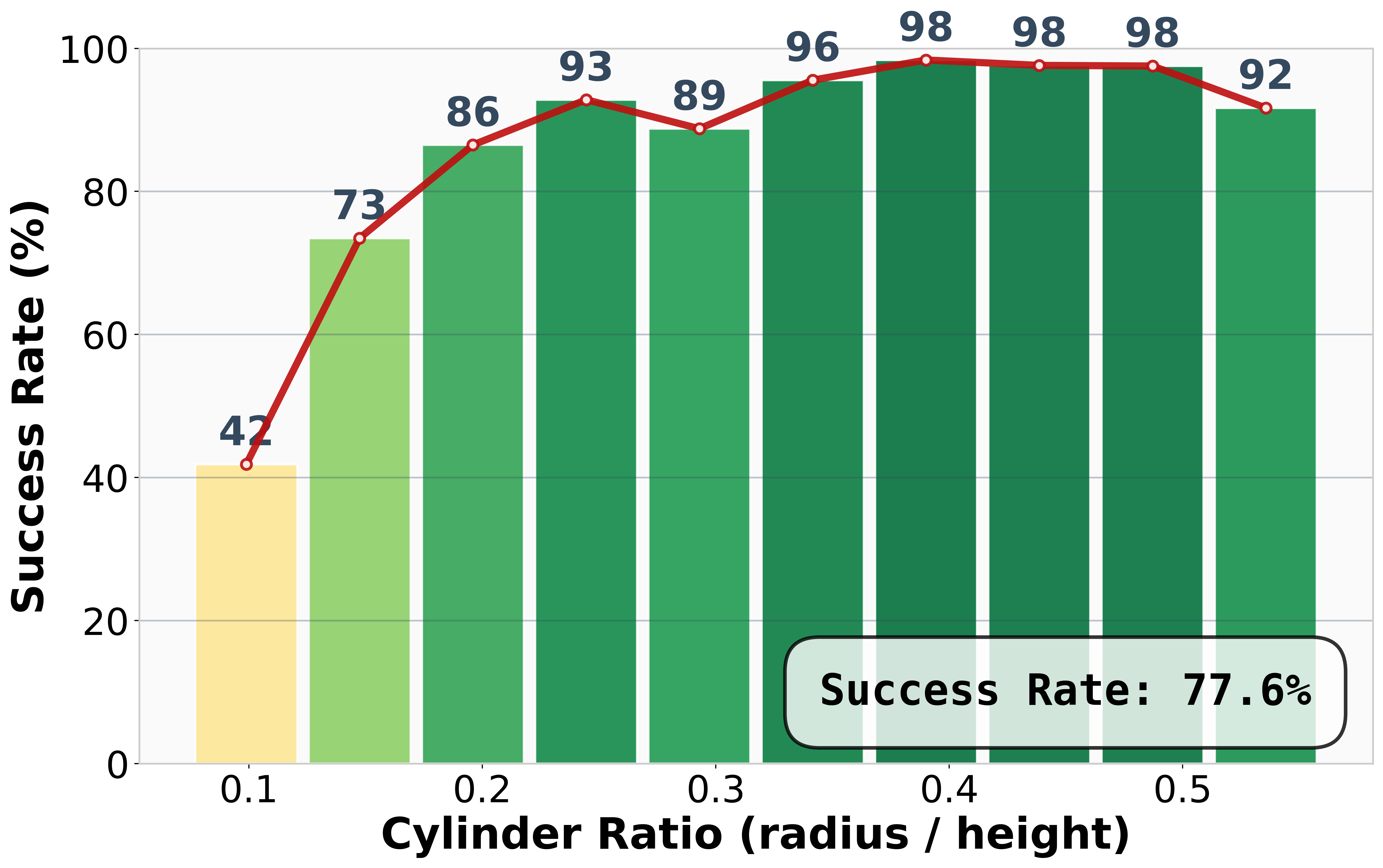}
  \caption{Success rate of \method across different object body scales in Push Robot task. }
  \label{fig:object_ratio}
  
\end{figure}

\subsection{Simulation Result}
\begin{figure*}[thpb]
  \centering
  \includegraphics[width=0.98\textwidth]{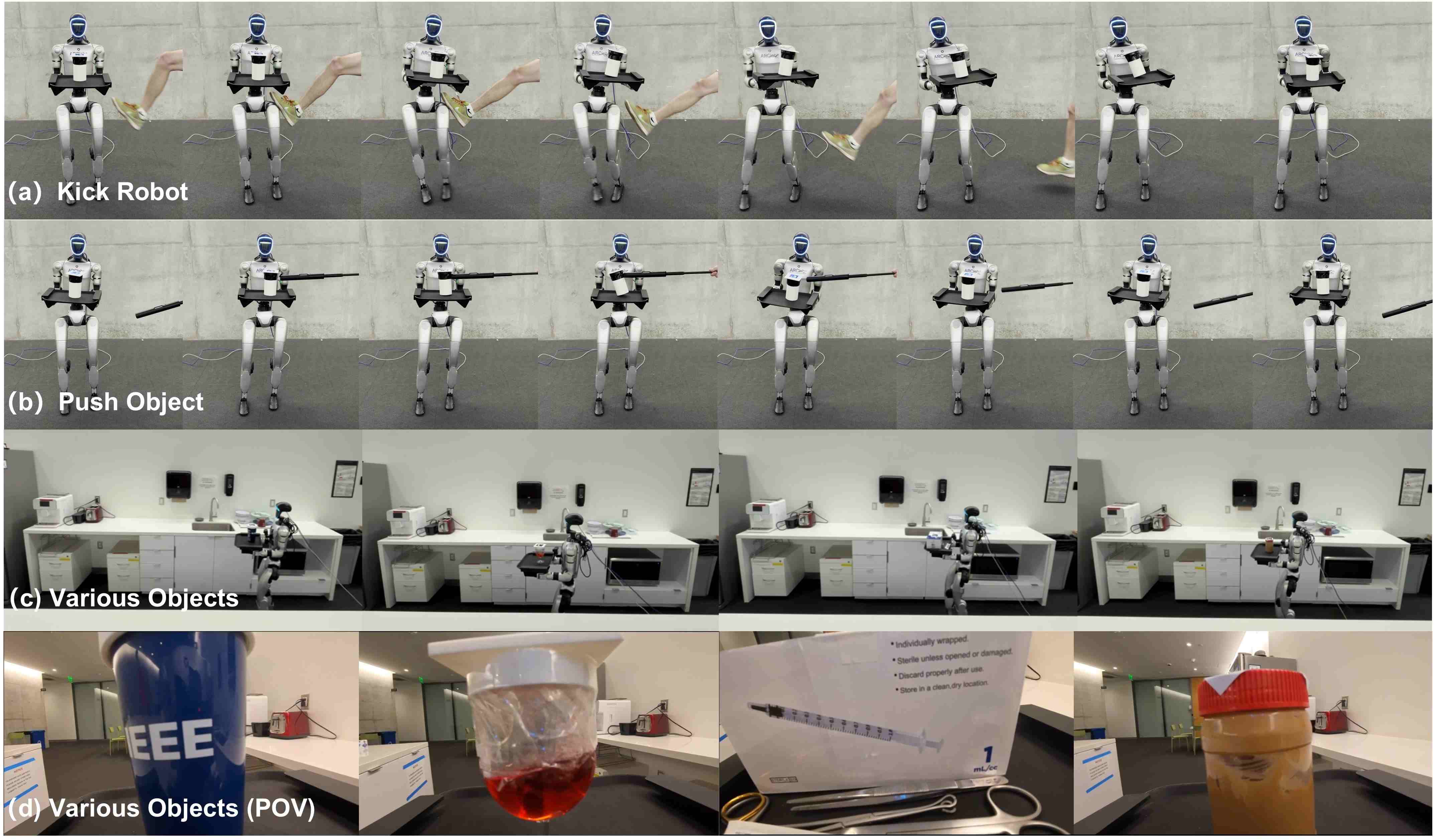}
  \caption{\method deployed on a Unitree G1 humanoid demonstrating payload stabilization capabilities. (\textbf{a}) Stabilizing the payload when the robot is kicked; (\textbf{b}) stabilizing the payload when the object is pushed; (\textbf{c},\textbf{d}) stably transporting various payloads.}
  \label{fig:real_world_exp}
\end{figure*}
To answer \textbf{Q2}, we evaluate \method in simulation under disturbances of varying types and magnitudes, and across objects with diverse scales.

Fig. \ref{fig:obj_robot_vel} demonstrates the velocity tracking and disturbance recovery behavior of \method under command switches and external pushes. 
The torso velocity closely tracks the commanded value while the object velocity remains tightly coupled to it. 
Under both robot-level pushes (10s, 13s) and object-level perturbation (around 16s), deviations are quickly corrected without sustained oscillation, demonstrating robust recovery. 

In Fig. \ref{fig:push_robot_polar} and \ref{fig:push_object_polar}, both plots illustrate \method maintains consistently high success rates across a wide range of push directions and magnitudes for both robot pushes and object pushes. 
As shown in Fig.~\ref{fig:push_polar}, \method maintains high success rates across diverse push directions and magnitudes for both robot and object pushes (74.4\% and 73.9\% respectively), with only moderate degradation under extreme forces, demonstrating robustness to both body-level and payload-level perturbations.

Fig.~\ref{fig:object_ratio} evaluates \method across varying cylinder radius-to-height ratios. Success rates remain consistently high across a broad range of aspect ratios, with only a modest drop for extremely slender objects (e.g., ratio 0.1), demonstrating robustness to diverse payload geometries.




\subsection{Real-World Results}
As shown in Fig.~\ref{fig:real_world_exp}, \method is deployed on a Unitree G1 humanoid robot to evaluate its payload stabilization capability under external disturbances and object variations in real-world environments.

Fig.~\ref{fig:real_world_exp}(a) and (b) show experiments where external disturbances are applied to the robot and the payload, respectively. 
In both cases, \method demonstrates timely whole-body recovery behaviors by coordinating upper and lower-body joints to re-stabilize the tray and prevent payload tipping under external disturbances. 
Recovery is achieved without abrupt corrective motions, demonstrating the robustness of the learned control strategy. 
Fig.~\ref{fig:real_world_exp}(c) and (d) demonstrate generalization to objects with diverse geometries and physical properties, including a standard coffee cup, a water-filled wine glass, medical and surgical tools, and a sealed food container. 
Despite their varying mass distributions, geometries, and contact properties, \method generalizes across all objects without retraining or fine-tuning. These results indicate strong sim-to-real transfer and robustness to real-world perception and dynamics.

\section{Discussion and Conclusion}
This paper formulates a new bimanual loco-manipulation problem, transporting unsecured objects on a tray with a bipedal humanoid robot under external disturbances, namely \task. This paper also proses \method, a residual student--teacher reinforcement learning framework for solving \task.
By decoupling locomotion from payload stabilization, the residual module resolves these competing objectives while preserving both gait quality and object stability.
Our training strategy further incorporates delayed object observations, control-latency, and domain randomization over physical and sensing parameters.
Together, these design choices reduce reliance on instantaneous measurements and promote temporally consistent control, improving sim-to-real transfer and robustness in real-world conditions.
Beyond the \task problem, \method may be extended to broader loco-manipulation skills that require task-specific sensing beyond standard proprioception, such as contact-rich manipulation during walking (e.g., door opening and cart pushing) and whole-body adaptation using vision or tactile feedback while preserving a robust pretrained gait.

Although potentially extensible, the current object encoding is limited to a single object and does not yet capture fine geometric or physical properties. 
Future work could integrate foundation pose-estimation models or vision-based RL to improve object representation.
Hardware constraints also limit perception: the fixed head-mounted camera has a narrow field of view.
This restricted viewpoint makes it difficult to transport objects in alternative holding poses, which are often needed in cluttered scenes to avoid collisions.
More broadly, pure sim-to-real RL often requires extensive reward shaping and careful simulator refinement, which becomes impractical for complex contact-rich loco-manipulation tasks.

\bibliographystyle{IEEEtran}
\bibliography{ref}

\clearpage
\appendix

\subsection{Observation Details}

\begin{table}[h]
  \centering
  \caption{Observation Terms}
  \label{tab:spaces_definitions}
  \begin{tabular}{lll}
    \toprule
    & Term & Dim \\
    \midrule
    Proprioception ($s^\text{prop}$)& Joint position & $\mathbb{R}^{29}$ \\
    & Joint velocity & $\mathbb{R}^{29}$ \\
    & Last action & $\mathbb{R}^{29}$ \\
    & Base angular velocity & $\mathbb{R}^{3}$\\
    & Base projected gravity & $\mathbb{R}^{3}$ \\
    \midrule
    Object ($s^\text{obj}$)      & Object position (camera frame) & $\mathbb{R}^{3}$ \\
    & Object quaternion (camera frame) & $\mathbb{R}^{4}$ \\
    \midrule
    Privilege ($s^\text{priv}$)   & Base linear velocity & $\mathbb{R}^{3}$ \\
    & Tray position (robot frame) & $\mathbb{R}^{3}$ \\
    & Tray projected gravity & $\mathbb{R}^{3}$ \\
    & Object position (tray frame) & $\mathbb{R}^{3}$ \\
    & Object angular velocity (tray frame) & $\mathbb{R}^{3}$ \\
    & Object linear velocity (tray frame) & $\mathbb{R}^{3}$ \\
    & Object projected gravity & $\mathbb{R}^{3}$ \\
    \midrule
    Goal ($g$)       & Velocity command & $\mathbb{R}^{3}$\\
    \bottomrule
  \end{tabular}
\end{table}

\subsection{Reward and Termination Details}
\label{section:appendix_reward}
All reward and termination terms used throughout training are summarized in TABLE~\ref{tab:reward_table}. The training procedure is divided into three stages. \textbf{Stage 1} (Pre-train Locomotion) trains a base locomotion policy with velocity tracking, body stability, and gait regularization rewards. \textbf{Stage 2} (Tray Holding Fine-tune) builds upon the \textbf{Stage 1} policy by introducing tray-specific rewards, including orientation, contact, and holder alignment, to learn stable tray transport. \textbf{Stage 3} (Residual Object Stabilization) trains the residual module with object-related rewards for keeping an object upright on the tray. Each subsequent stage inherits all terms from the previous stage, modifying or disabling specific terms as needed.

For task-specific failure conditions such as the tray or object dropping, we employ a \textit{delayed-timeout} mechanism rather than immediate episode termination (marked with $^\dagger$ in TABLE \ref{tab:reward_table}), where episodes are truncated when a tracked condition remains violated for a configured duration (1.0 s in our setup). This design serves two purposes. First, it improves sampling efficiency as continued simulation yields minimal task-relevant learning signal once the object or tray is unrecoverable. Second, the delay window allows the policy to observe post-failure states and learn to maintain stable locomotion, which is critical for safe real-world deployment where unseen object states may trigger erratic behavior. In addition, the value function bootstrap is preserved in truncated episodes and the critic does not incorrectly learn to associate post-failure states with zero future return.

\begin{table*}[htbp]
\centering
\caption{Reward and termination terms across three training stages. 
Terms marked with $^\dagger$ are \emph{delayed-timeout} termination conditions: they do not immediately reset the episode, but will trigger a timeout (truncation) after the configured delay has elapsed. 
Terms marked with $^\ddagger$ indicate a change from the previous stage.}
\label{tab:reward_table}
\footnotesize
\setlength{\tabcolsep}{4pt}
\renewcommand{\arraystretch}{1.15}
\begin{tabular}{@{} ll @{\hspace{25pt}} ll @{}}
\toprule
\textbf{Training Stage} & \textbf{Term} & \textbf{Expression} & \textbf{Weight} \\
\midrule


\multirow{21}{2.8cm}{\textbf{Stage 1:}\\Pre-train\\Locomotion}
& \multicolumn{3}{@{}l}{\textit{Locomotion tracking rewards}} \\
& Track linear vel
  & $\exp\!\bigl(-\big\|\mathbf{v}^{r_{x,y}}_t -\mathbf{\hat{v}}^{r_{x,y}}_t\big\|^2_{2} / \sigma^2\bigr)$,\; $\sigma{=}\sqrt{0.25}$
  & 1.0 \\
& Track angular vel
  & $\exp\!\bigl(-\big\|\boldsymbol{\omega}^{r_\text{yaw}}_t -\boldsymbol{\hat{\omega}}^{r_\text{yaw}}_t\big\|^2_{2} / \sigma^2\bigr)$,\; $\sigma{=}\sqrt{0.25}$
  & 1.5 \\[2pt]

& \multicolumn{3}{@{}l}{\textit{Survival \& body stability rewards}} \\
& Robot Alive bonus
  & $\mathbf{1}_{\text{[alive]}}$
  & 1.0 \\
& Torso vertical lin vel
  & $\exp\!\bigl(-\lambda\, v^{r_z\ 2}_t\bigr)$,\; $\lambda{=}2$
  & 0.2 \\
& Torso roll/pitch ang vel
  & $\exp\!\bigl(-\lambda\,\big\|\boldsymbol{\omega}^{r_{x,y}}_t \big\|^2_{2}\bigr)$,\; $\lambda{=}1$
  & 0.2 \\
& Upright torso
  & $\exp(-\lambda \big\|\mathbf{P}_{xy}\big(\mathbf{g}^\text{r}_t)\big\|^2_{2})$,\; $\lambda{=}4$
  & 0.25 \\
& Torso height
  & $\exp\!\bigl(-\lambda\,|h_{\mathrm{torso}} - h_{\mathrm{target}}|\bigr)$,\; $\lambda{=}10$, $h_{\mathrm{target}}{=}0.82$
  & 0.5 \\[2pt]

& \multicolumn{3}{@{}l}{\textit{Regularization penalties}} \\
& Joint velocity
  & $\sum_i \dot{q}_{i,t}^2$
  & $-$0.001 \\
& Joint acceleration
  & $\sum_i \ddot{q}_{i,t}^2$
  & $-2.5{\times}10^{-7}$ \\
& Action rate
  & $\sum_i (a_{i,t} - a_{i,t-1})^2$
  & $-$0.05 \\
& Joint position limits
  & $\sum_i \bigl[\max(0, q_{i,t}{-}q_{i,t}^{u}) + \max(0, q_{i,t}^{l}{-}q_{i,t})\bigr]$
  & $-$5.0 \\
& Energy consumption
  & $\sum_i |\dot{q}_{i,t}|\,|\tau_{i,t}|$
  & $-2{\times}10^{-5}$ \\
& Arm joint deviation (L1)
  & $\sum_{i\in\mathrm{arms}} |q_{i,t} - q_{i,t}^{\mathrm{default}}|$
  & $-$0.5 \\
& Waist yaw joint deviation
  & $|q_{\mathrm{waist\_yaw},t} - q_{\mathrm{waist\_yaw},t}^{\mathrm{default}}|$
  & $-$1.0 \\[2pt]

& \multicolumn{3}{@{}l}{\textit{Feet rewards \& penalties}} \\
& Feet slide
  & $\sum_k \lVert\mathbf{v}_{\mathrm{foot}_k,t}^{xy}\rVert \cdot \mathbf{1}_{[\mathrm{foot}_k\text{\_contact}]}$
  & $-$0.2 \\
& Foot clearance
  & $\exp\!\bigl(-\frac{1}{\sigma}\sum_k (z_{\mathrm{foot}_k, t} - 0.1)^2 \cdot \tanh(2\lVert\mathbf{v}^{xy}_{\mathrm{foot}_k, t}\rVert)\bigr)$,\; $\sigma{=}0.05$
  & 1.0 \\
& Feet impact
  & $\exp\!\bigl(-\lambda \left\|\mathbf{v}^{\text{feet}}_{z,t}-\mathbf{v}^{\text{feet}}_{z,t-1}\right\|_2^2\bigr)$,\; $\lambda{=}5$
  & 0.5 \\
& Undesired contacts
  & $\mathbf{1}_{\text{[undesired\_body\_contact]}}$
  & $-$1.0 \\[2pt]

& \multicolumn{3}{@{}l}{\textit{Termination conditions}} \\
& Time out
  & Episode length exceeded ($T{=}20\,$s)
  & --- \\
& Robot Base height
  & $\mathbf{1}_{[z^r_t < 0.4]}$ (terminated)
  & --- \\
& Robot Bad orientation
  & $\mathbf{1}_{[\mathrm{tilt}_{\mathrm{torso}} > 0.7\,\mathrm{rad}]}$ (terminated)
  & --- \\

\midrule


\multirow{14}{2.8cm}{\textbf{Stage 2:}\\Pre-train\\Tray Holding\\Fine-tune}
& \multicolumn{3}{@{}l}{\textit{Inherits all Stage~1 terms, with the following changes and additions:}} \\[2pt]

& \multicolumn{3}{@{}l}{\textit{Modified terms from Stage~1}$^\ddagger$} \\
& Arm joint deviation (exp)$^\ddagger$
  & $\exp\!\bigl(-\lambda \sum_{i\in\mathrm{arms}} |q_{i,t} - q_{i,t}^{\mathrm{default}}|\bigr)$,\; $\lambda{=}0.3$
  & 0.5 \\[2pt]

& \multicolumn{3}{@{}l}{\textit{Tray stability rewards (new)}} \\
& Tray upright
  & $\exp(-\lambda \big\|\mathbf{P}_{xy}\big(\mathbf{g}^\text{tray}_t)\big\|^2_{2})$,\; $\lambda{=}4$
  & 0.25 \\
& Tray vertical lin vel
  & $\exp\!\bigl(-\lambda\, v_{z,t}^{\mathrm{tray}\,2}\bigr)$,\; $\lambda{=}2$
  & 0.2 \\
& Tray roll/pitch ang vel
  & $\exp\!\bigl(-\lambda\,\big\|\boldsymbol{\omega}^{\text{tray}_{x,y}}_t \big\|^2_{2}\bigr)$,\; $\lambda{=}1$
  & 0.2 \\
& Tray-EE contact
  & $\mathbf{1}_{[\text{tray\_ee\_contact}]}$
  & 0.01 \\
& Tray-EE contact force
  & $\exp\!\bigl(-\lambda \big\|\boldsymbol{F}^{\text{tray\_ee}}_t \big\|_{2}\bigr)$,\; $\lambda{=}0.005$
  & 0.2 \\
& Tray--Left-EE quat alignment
  & $\exp\!\bigl(-\lambda \cdot 2\arccos(|\mathbf{q}_t^{\mathrm{tray}} \cdot \mathbf{q}_t^{\mathrm{left\_ee}}|)^2\bigr)$,\; $\lambda{=}2$
  & 0.5 \\
& Tray--Left-EE quat alignment
  & $\exp\!\bigl(-\lambda \cdot 2\arccos(|\mathbf{q}_t^{\mathrm{tray}} \cdot \mathbf{q}_t^{\mathrm{right\_ee}}|)^2\bigr)$,\; $\lambda{=}2$
  & 0.5 \\
& Joint torque penalty
  & $\sum_i \tau_{i,t}^2$
  & $-2{\times}10^{-5}$ \\[2pt]

& \multicolumn{3}{@{}l}{\textit{Termination conditions (new)}} \\
& Tray fallen$^\dagger$
  & $\mathbf{1}_{[z_{\mathrm{tray}} < 0.7]}$ (timeout after $1.0\,$s)
  & --- \\

\midrule


\multirow{10}{2.8cm}{\textbf{Stage 3:}\\Residual Object\\Stabilization}
& \multicolumn{3}{@{}l}{\textit{Inherits all Stage~2 terms, with the following changes and additions:}} \\[2pt]

& \multicolumn{3}{@{}l}{\textit{Disabled terms from Stage~2}$^\ddagger$} \\
& Tray vertical lin vel$^\ddagger$
  & (disabled)
  & --- \\
& Tray roll/pitch ang vel$^\ddagger$
  & (disabled)
  & --- \\[2pt]

& \multicolumn{3}{@{}l}{\textit{Object stability rewards (new)}} \\
& Object-tray contact
  & $\mathbf{1}_{\{\text{object\_tray\_contact}\}}$
  & 0.05 \\
& Object upright
  & $\exp(-\lambda \big\|\mathbf{P}_{xy}\big(\mathbf{g}^\text{obj}_t)\big\|^2_{2})$,\; $\lambda{=}4$
  & 1.0 \\[2pt]

& \multicolumn{3}{@{}l}{\textit{Termination conditions (new)}} \\
& Object fallen$^\dagger$
  & $\mathbf{1}_{[z_{\mathrm{obj}} < 0.7]}$ (timeout after $1.0\,$s)
  & --- \\
& Object bad orientation$^\dagger$
  & $\mathbf{1}_{[\mathrm{tilt}_{\mathrm{obj}} > 0.7\,\mathrm{rad}]}$ (timeout after $1.0\,$s)
  & --- \\

\bottomrule
\end{tabular}
\end{table*}

\subsection{Domain Randomization}

All domain randomization terms used throughout training are summarized in TABLE~\ref{tab:domain_randomization}.

\begin{table}[htbp]
\centering
\caption{Domain randomization parameters}
\label{tab:domain_randomization}
\footnotesize
\setlength{\tabcolsep}{4pt}
\renewcommand{\arraystretch}{1.15}
\begin{tabular}{@{} ll @{}}
\toprule
\textbf{Parameter} & \textbf{Range / Value} \\
\midrule
Robot static friction          & $[0.3, 1.0]$ \\
Robot dynamic friction         & $[0.3, 1.0]$ \\
End-effector static friction         & $[2.0, 3.0]$ \\
End-effector dynamic friction        & $[1.5, 2.5]$ \\
End-effector restitution             & $[0.0, 0.05]$ \\
End-effector mass                    & $[0.05, 0.3]\,$kg \\
Torso added mass             & $[-2.0, 3.0]\,$kg \\
Torso CoM offset             & $x{\in}[\pm0.025]$, $y{\in}[\pm0.05]$, $z{\in}[\pm0.05]\,$m \\
Tray static friction                & $[1.2, 2.0]$ \\
Tray dynamic friction               & $[1.0, 1.8]$ \\
Tray restitution                    & $[0.0, 0.05]$ \\
Tray mass                           & $[0.3, 0.7]\,$kg \\
Object radius scale                 & $[0.7, 1.5]\times$0.03m \\
Object height scale                 & $[0.75, 2.0]\times$0.1m \\
Object static friction              & $[0.5, 1.0]$ \\
Object dynamic friction             & $[0.4, 0.9]$ \\
Object restitution                  & $[0.0, 0.5]$ \\
Object mass                         & $[0.05, 1.0]\,$kg \\
Object CoM offset                   & $x{\in}[\pm0.01]$, $y{\in}[\pm0.01]$, $z{\in}[\pm0.02]\,$m \\
Joint velocity reset                & $[-1.0, 1.0]\,$rad/s \\
External push on robot              & every $[3, 5]\,$s, $v_{xy}{\in}[\pm0.5]\,$m/s \\
External push on object             & every $[2, 4]\,$s,  \\
                                    & $v_{xy}{\in}[\pm0.3]\,$m/s, $\omega_{xy}{\in}[\pm0.3]\,$rad/s\\
\bottomrule
\end{tabular}
\end{table}

\subsection{Training Hyperparameter}

TABLE~\ref{tab:teacher_hyperparams} and TABLE~\ref{tab:student_hyperparams} summarize the training hyperparameters for the residual teacher and encoder distillation stages, respectively.
We experiment with two residual designs: residual FiLM and residual action, both of which share the same Transformer-based history encoder.
The encoder distillation stage inherits the full teacher architecture and trains a student encoder to replicate the teacher's embeddings from proprioceptive and object observations.
All experiments are conducted on two NVIDIA RTX A6000 GPUs with 4\,096 parallel environments per GPU.

\begin{table}[htbp]
\centering
\caption{Hyperparameters for residual teacher policy (PPO). Both adapter variants share the same base MLP, encoder, and PPO configuration.}
\label{tab:teacher_hyperparams}
\footnotesize
\renewcommand{\arraystretch}{1.15}
\begin{tabular}{@{} l l @{}}
\toprule
\textbf{Hyperparameters} & \textbf{Values} \\
\midrule
\multicolumn{2}{@{}l}{\textit{Runner}} \\
Number of environments           & 8\,192 (4\,096 $\times$ 2 GPUs) \\
Number of steps per environment  & 24 \\
\midrule
\multicolumn{2}{@{}l}{\textit{Base policy}} \\
MLP size (actor \& critic)       & {[}512, 256, 128{]} \\
\midrule
\multicolumn{2}{@{}l}{\textit{Transformer encoder}} \\
Context dimension ($d_{\mathrm{ctx}}$) & 64 \\
Encoder layers                   & 2 \\
Attention heads                  & 4 \\
\midrule
\multicolumn{2}{@{}l}{\textit{Residual FiLM adapter}} \\
FiLM hidden dim               & 128 \\
$\gamma$ clamp                   & $\pm 3.0$ \\
\midrule
\multicolumn{2}{@{}l}{\textit{Residual action adapter}} \\
Residual MLP size                & {[}512, 256, 128{]} \\
\midrule
\multicolumn{2}{@{}l}{\textit{Algorithm (PPO)}} \\
LR schedule                      & Adaptive \\
Discount factor ($\gamma$)       & 0.99 \\
GAE $\lambda$                    & 0.95 \\
Clip parameter ($\epsilon$)      & 0.2 \\
Value loss coefficient           & 1.0 \\
Entropy coefficient              & 0.01 \\
Desired KL                       & 0.01 \\
Num learning epochs              & 5 \\
Num mini-batches                 & 4 \\
Max gradient norm                & 1.0 \\
\bottomrule
\end{tabular}
\end{table}

\begin{table}[htbp]
\centering
\caption{Hyperparameters for encoder distillation.}
\label{tab:student_hyperparams}
\footnotesize
\renewcommand{\arraystretch}{1.15}
\begin{tabular}{@{} l l @{}}
\toprule
\textbf{Hyperparameters} & \textbf{Values} \\
\midrule
\multicolumn{2}{@{}l}{\textit{Runner}} \\
Number of environments           & 8\,192 (4\,096 $\times$ 2 GPUs) \\
Number of steps per environment  & 24 \\
\midrule
\multicolumn{2}{@{}l}{\textit{Algorithm (Distillation)}} \\
Learning rate                    & $5 \times 10^{-5}$ \\
Num learning epochs              & 1 \\
Max gradient norm                & 0.5 \\
Loss type                        & MSE \\
\bottomrule
\end{tabular}
\end{table}

\subsection{Training Strategy}

In this section, we detail an additional strategy applied during Stage~1 (pre-training locomotion)
to accelerate policy training.

In whole-body policy training, the large action space can significantly slow convergence,
as the optimizer simultaneously explores locomotion gaits and upper-body movements.
To address this, we introduce an auxiliary \emph{upper-body curriculum} loss during Stage~1
that effectively freezes the upper-body joints, allowing the policy to focus exclusively
on learning stable lower-body locomotion.

Concretely, we add a regularization term to the PPO loss that penalizes both the
mean and standard deviation of the policy's action distribution on specified upper-body
joint indices $\mathcal{J}_{\mathrm{upper}}$:
\begin{equation}
    \mathcal{L}_{\mathrm{upper}} = \alpha \left(
        \frac{\sum_{b,j} \mu_{b,j}^2}{|\mathcal{B}||\mathcal{J}_{\mathrm{upper}}|}
        + \frac{1}{2} \cdot
        \frac{\sum_{b,j} \sigma_{b,j}^2}{|\mathcal{B}||\mathcal{J}_{\mathrm{upper}}|}
    \right)
\end{equation}
where $\mu_{b,j}$ and $\sigma_{b,j}$ are the mean and standard deviation of the
policy's action distribution for joint $j$ in sample $b$, $\mathcal{B}$ denotes
the mini-batch, and $\alpha$ is a coefficient that can be decayed over training
via a curriculum schedule.

The L2 penalty on $\mu$ drives the upper-body actions toward zero (i.e., the default
pose), while the penalty on $\sigma$ suppresses exploration on these joints, preventing
the optimizer from wasting gradient updates on uninformative upper-body movements during
early locomotion learning. This regularizer is applied only in Stage~1 and removed in
subsequent stages, where the upper-body joints are freed to learn tray-holding and
object-stabilization behaviors. In practice, this strategy reduces Stage 1 training time substantially by suppressing exploration on upper-body joints and focusing gradient updates on the lower-body joints relevant for locomotion.
\end{document}